\documentclass[letterpaper, 10 pt, conference]{ieeeconf}  %
\IEEEoverridecommandlockouts      

\usepackage{amsmath} 
\usepackage{amssymb}  
\usepackage{graphicx}
\usepackage{multirow}
\usepackage{array}
\usepackage{tabularx}
\usepackage{xcolor}                        

\title{\LARGE \bf MLLM-SUL: Multimodal Large Language Model for Semantic Scene Understanding and Localization in Traffic Scenarios}

\author{Jiaqi Fan$^{1}$\thanks{$^{1}$Jiaqi Fan is with is with Shanghai Research Institute for Intelligent Autonomous Systems, Tongji University, Shanghai 201804, China (e-mail: fanjq@tongji.edu.cn).}, 
Jianhua Wu$^{2}$, Jincheng Gao$^{2}$, Jianhao Yu$^{2}$\thanks{$^{2}$Jianhua Wu, Jincheng Gao, Jianhao Yu, Hongqing Chu and Bingzhao Gao are with the School of Automotive Studies, Tongji University, Shanghai 201804, China (e-mail: 2332980@tongji.edu.cn, 1953115@tongji.edu.cn, 1953394@tongji.edu.cn, chuhongqing@tongji.edu.cn, gaobz@tongji.edu.cn). \emph{(Corresponding author: Hongqing Chu)}}, 
Yafei Wang$^{3}$\thanks{$^{3}$Yafei Wang  is with the School of Mechanical Engineering, Shanghai Jiao Tong University, Shanghai, 200240, China (e-mail: wyfjlu@sjtu.edu.cn).}, 
Hongqing Chu$^{2}$, and Bingzhao Gao$^{2}$   
}

\begin{document}

\maketitle
\thispagestyle{empty}
\pagestyle{empty}

\begin{abstract}
	Multimodal large language models (MLLMs) have shown satisfactory effects in many autonomous driving tasks. In this paper, MLLMs are utilized to solve joint semantic scene understanding and risk localization tasks, while only relying on front-view images. In the proposed MLLM-SUL framework, a dual-branch visual encoder is first designed to extract features from two resolutions, and rich visual information is conducive to the language model describing risk objects of different sizes accurately. Then for the language generation, LLaMA model is fine-tuned to predict scene descriptions, containing the type of driving scenario, actions of risk objects, and driving intentions and suggestions of ego-vehicle. Ultimately, a transformer-based network incorporating a regression token is trained to locate the risk objects. Extensive experiments on the existing DRAMA-ROLISP dataset and the extended DRAMA-SRIS dataset demonstrate that our method is efficient, surpassing many state-of-the-art image-based and video-based methods. Specifically, our method achieves 80.1\% BLEU-1 score and 298.5\% CIDEr score in the scene understanding task, and 59.6\% accuracy in the localization task. Codes and datasets are available at https://github.com/fjq-tongji/MLLM-SUL.
	
	
	

\end{abstract}

\section{INTRODUCTION}

Recently, using natural language as a unified output for various automation tasks has attracted the attention of many researchers~\cite{LLM_autonomous_TITS,LLM_autonomous_IROS_2024_1,LLM_autonomous_IROS_2024_2,LLM_autonomous_RAL}. Compared with conventional optimization-based methods, language-based models have shown remarkable performances, particularly in interpretability and generation. These methods typically utilize a structured language as text prompts including driving rules and scenario descriptions. Then, the prompts are input to large language models (LLMs), such as LLaMA~\cite{llama3} and GPT-4o~\cite{GPT-4}, to generate driving suggestions. In~\cite{LLM_autonomous_drivr_like_a_human} and~\cite{LLM_autonomous_TrafficGPT}, GPT is utilized to generate high-level driving instructions, showing the excellent reasoning abilities of LLM. These works have confirmed the feasibility and great potential of applying language models for traffic scenarios. 

Correspondingly, many language datasets~\cite{EMMA,DriveVLM} specific to traffic scenarios have been built in recent years to meet the training or fine-tuning needs of LLMs. For example, IDD-X dataset~\cite{IDDX} contains the localization and explanation of critical objects in the surrounding driving scene, and Rank2tell dataset~\cite{Rank2tell} provides importance level ranking and linguistic descriptions of several risk objects in the driving scenes. These datasets typically only involve textual annotations for images, without numerical explanations. Especially, Malla~\emph{et al.}~\cite{drama} establish DRAMA dataset, which can describe risk objects in current driving scenes while providing the coordinates of corresponding objects. In our work, the DRAMA dataset is further expanded to include more diverse information including the category of the current traffic scene.


Furthermore, some researchers explore encoding visual inputs such as images or videos into LLMs, enabling the language model to understand non-textual contents. To bridge the gap between visual and textual modalities, these multimodal large language models (MLLMs) generally contain an additional sub-network to fit different inputs. In BLIP-2~\cite{blip2} and InstructBLIP~\cite{InstructBLIP}, several learnable queries are performed to learn the interactions between images and texts through self-attention and cross-attention operations. In LLaMA-Adapter~\cite{llama-adapter}, the visual information is incorporated into the lightweight adaption layer and fine-tuned with the language model through attention networks. Thus, these image-text alignment methods make multimodal scene understanding possible in traffic scenarios.

\begin{figure}[!t]
	\centering
	\includegraphics[width=1.0\linewidth]{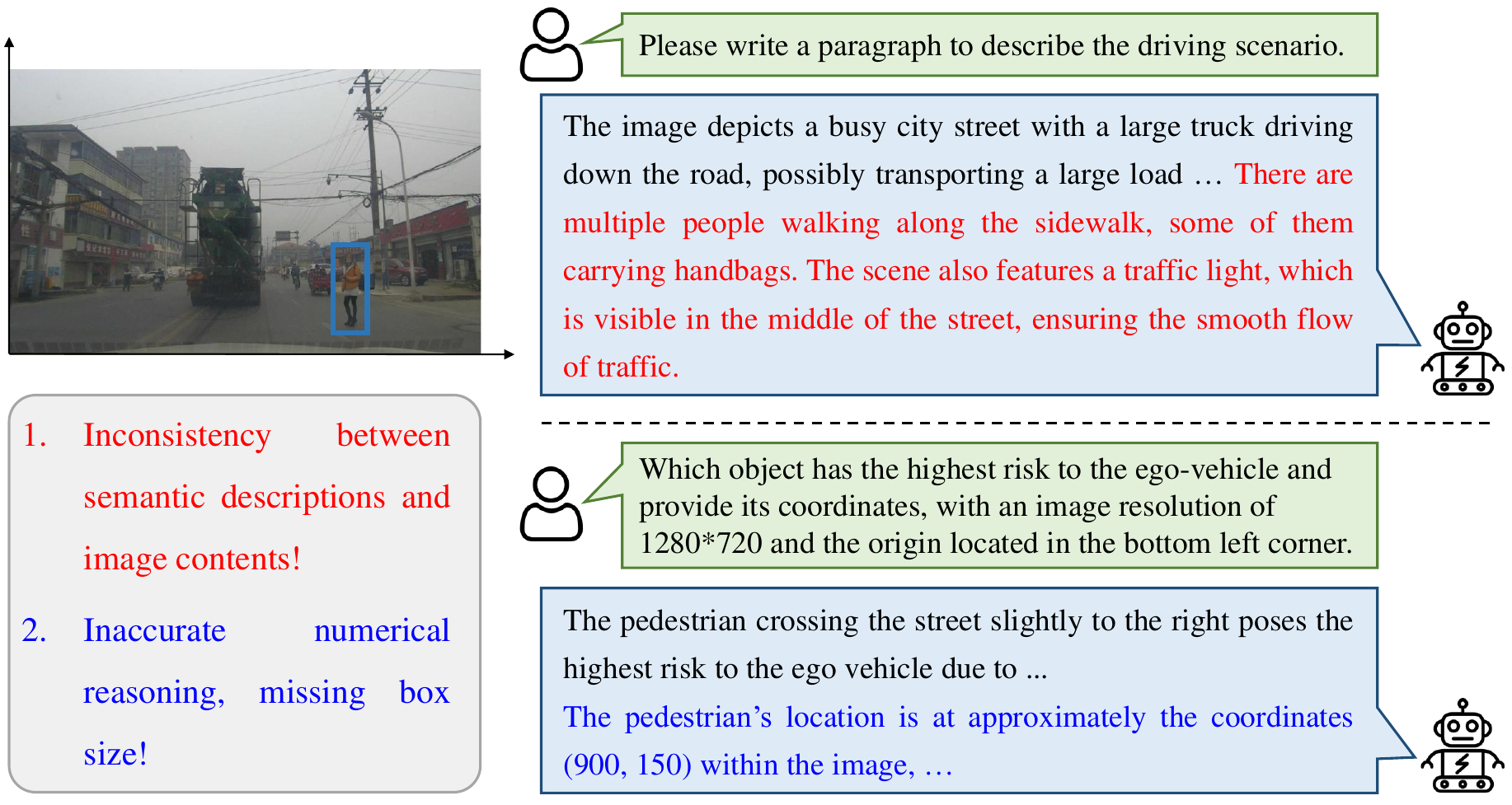}
	\caption{An example of errors in MLLM. Wrong semantic descriptions and numbers are highlighted in red and blue, respectively.}
	\label{Fig_intro}
\end{figure}

\textcolor{black}{However, as shown in Fig.~\ref{Fig_intro}, there are still two key problems in MLLMs: 1) a weak visual model can cause information loss during the feature encoding, resulting in inconsistency between semantic descriptions and image contents; 2) the reasoning ability of language models for numbers is not as strong as that for texts.} To address two problems, we propose a novel multimodal large model named MLLM-SUL based on image inputs to accomplish joint semantic scene understanding and localization tasks in traffic scenarios. \textcolor{black}{Regarding the first issue, a dual-branch visual encoder is built in this paper, combining grid features from the backbone network and region features from the detector. This design can fully extract the perceptual information contained in images, which is beneficial for guiding language models to generate more accurate descriptions.} \textcolor{black}{Regarding the second issue, a query-based transformer network is added to regress the coordinates of risk objects instead of predicting coordinates from the language model directly. This structure avoids the problem of inaccurate prediction of numbers by large language models.} Generally speaking, video-based methods are more conducive to scene understanding than image-based methods. Despite this gap, our method still achieves a 1.6\% improvement in BLEU-4 score, 2.5\% improvement in METEOR score, and an average 0.8\% improvement in two tasks than the state-of-the-art video-based methods on the DRAMA-ROLISP and DRAMA-SRIS dataset.

To summarize, the contributions of this paper are summarized as follows:
\begin{itemize}
	\item We propose an image-based unified multimodal architecture MLLM-SUL for scene understanding of traffic scenarios. In our method, an end-to-end large model that aggregates multiple visual components is built, capable of accomplishing two driving tasks simultaneously: semantic scene description and risk object localization.

	\item The proposed framework includes a dual-branch visual encoder to extract region and grid features from low- and high-resolution images, which guide the language model to generate accurate texts. After obtaining the descriptions, a query-based regression network is designed to predict the localization of risk objects. 
	
	
	\item We conduct extensive experiments on the public DRAMA-ROLISP dataset and extended DRAMA-SRIS dataset. The results demonstrate that the proposed MLLM-SUL outperforms many state-of-the-art multimodal language models, including BLIP-2, InstructBLIP, and LLaMA-Adapter.
\end{itemize}


\section{RELATED WORK}

\subsection{\textcolor{black}{Semantic Scene description methods}}
Previous image captioning methods include two types: CNN-LSTM methods~\cite{Caption_CNN_LSTM_1,Caption_CNN_LSTM_2} and transformer-based methods~\cite{BENet,RCMF}. The former adopts a convolutional neural network (CNN) to extract image features and employs a long short-term memory (LSTM) network to decode and obtain text expressions. This structure is easy to implement, but not conducive to finding the correlations between visual and textual modalities. The latter uses a self-attention mechanism for global image encoding and sequence modeling. \textcolor{black}{DASTNet~\cite{DASTNet} presents an adaptive shared transformer block and a gating mechanism to adaptively calculate the importance of features, preserving stronger visual inputs.} \textcolor{black}{DAIT~\cite{DAIT} devises a dual-adaptive interactive transformer network for captioning, realizing an automatic semantic decoding process.} Due to the excellent generation ability of the attention model, transformer-based methods are widely applied in caption generation tasks in many fields.


In recent years, many MLLMs have emerged to solve captioning tasks. These methods typically contain a visual encoder, a vision-language alignment module, and a language model for text generation. LLaVA~\cite{LLaVA} connects the visual encoder CLIP and language decoder Vicuna for instruction fine-tuning, and the trained end-to-end model can generate conversations, detailed descriptions, and complex reasoning. Video-LLaMA~\cite{video-llama} applies a cross-modal language generation method for video comprehension, in which a video Q-former and an audio Q-former module are utilized to aggregate frame-level representations into the language generator. In eP-ALM~\cite{ep-alm}, the perceptual and textual inputs are fed to ViT and large language model, respectively, and the predicted descriptions are consistent with visual contents. Different from these methods, the visual encoder of our model provides stronger inputs for the language model by building two branches with different resolutions, enhancing the text reasoning ability.


\subsection{Multimodal understanding in traffic scenarios}
Recently, many researchers have explored integrating MLLMs into an end-to-end framework to complete multiple scene understanding tasks in autonomous driving. \textcolor{black}{VLAAD~\cite{VLAAD} aggregates video frames of driving scenes into the LLaMA model to complete complex description and reasoning tasks. In the generated answer, the driving intention and control action of ego-vehicle are are described in detail.} 
\textcolor{black}{Dolphins~\cite{Dolphins} uses image or video data as input for the open-sourced vision-language model and further completes comprehensive scene understanding tasks, including prediction, instant adaptation, error recovery, conversation, etc.}
DriveGPT4~\cite{MLLM_autonomous_drivegpt4} takes the encoded multi-frame video sequences as inputs into the GPT-4 model to predict the next control signals, such as vehicle speed and steering angle. 

In BEV-InMLLM~\cite{MLLM_autonomous_holistic}, multi-view videos are processed with a frozen language model to generate driving responses, in which a Q-Former is introduced to inject bird's-eye-view representations into the language model. In RAG-Driver~\cite{MLLM_autonomous_rag_driver}, video embeddings, control signals, and task instructions are input to the LLaMA model to explain the current actions and predict the next action signals. In HiLM-D~\cite{HiLM-D}, the low-resolution video and high-resolution image frame are together input to a unified language-based paradigm, generating descriptions about risk objects and ego-vehicle actions. Unlike these methods, our model does not directly infer numbers from language models, reducing the difficulty of text prediction.

\begin{figure*}[!t]
	\centering
	\includegraphics[width=1.0\linewidth]{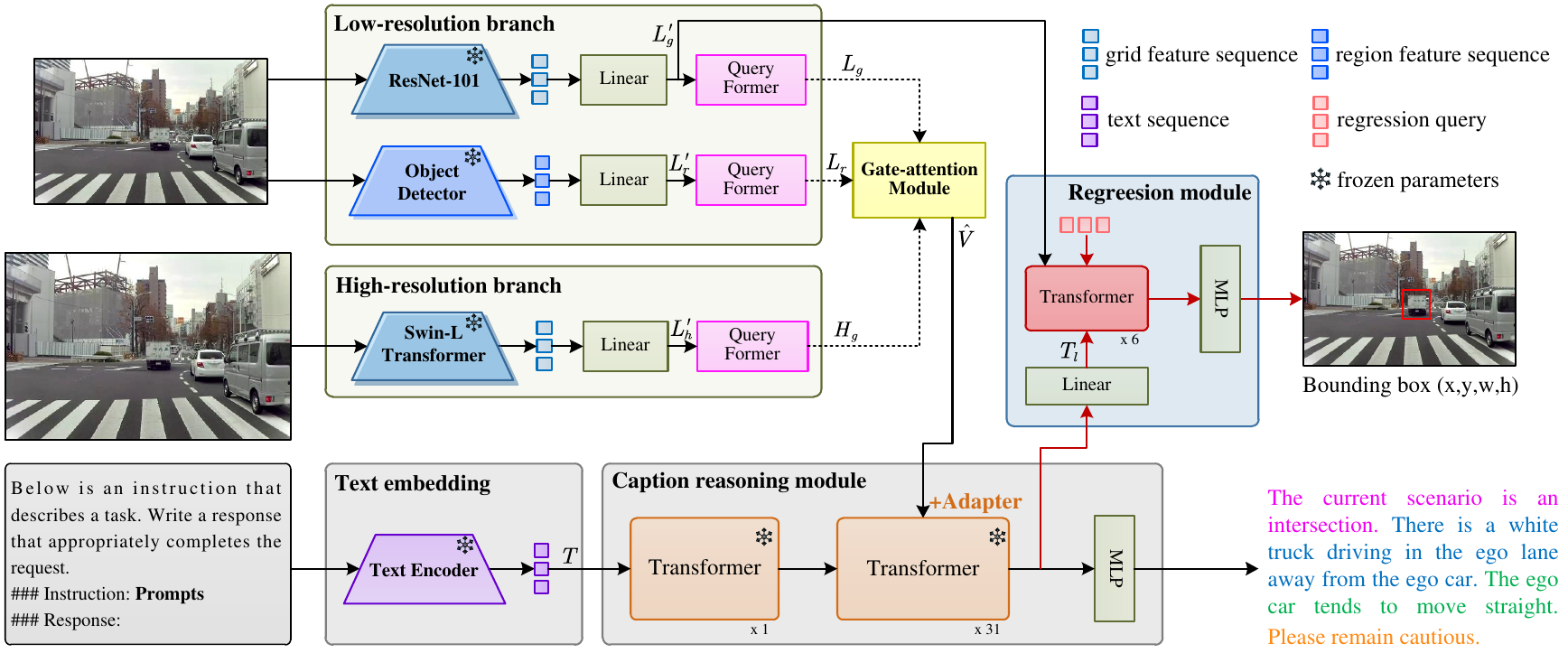}
	\caption{Overall structure of the proposed MLLM-SUL model. The prompts are: \emph{What is the current driving scenario? Which object is at the highest risk? Then predict the intentions and suggestions for the ego-car.}}
	\label{Fig_overall}
\end{figure*}

\section{METHOD}
As shown in Fig.~\ref{Fig_overall}, the proposed MLLM-SUL model consists of three main components: a dual-branch visual encoder to extract and fuse features from low-resolution and high-resolution images (Sec.~\ref{Dual-branch visual feature extractor}), a LLaMA-based language model to reason descriptions for driving scenario, risk object and ego-vehicle (Sec.~\ref{caption reasoning module}), and a regression module to locate the risk object in the image (Sec.~\ref{localization module}). Three modules are closely integrated to complete the caption generation and detection task jointly.

\subsection{Dual-branch visual encoder}
\label{Dual-branch visual feature extractor}
\subsubsection{Low-resolution branch}
In our model, a frozen visual backbone network and object detector are utilized to extract grid and region features to better learn the contextual representations. Here, ResNet-101~\cite{Resnet} is selected as the backbone network due to its powerful feature extraction ability, as evaluated by many previous works. For the input image with the resolution of $H$$\times$$W$, it is downsampled 32 times, obtaining $H/32$$\times$$W/32$ output. Then, the obtained feature map is flattened to a grid feature sequence, and a linear projection is performed to maintain a fixed output dimension. To further bridge the gap between the visual encoder and language generator, we train a query former with a fixed query length to restrict the number of output features.

The object-level features are extracted from the frozen detector Faster R-CNN~\cite{fasterrcnn}, which is pre-trained on the Visual Genome dataset. In this detector, each image is detected with several proposals, with a minimum number of 10 and a maximum number of 100. And in the proposed structure, a linear projection layer and a query former are utilized after the detector to better unify the dimensions and number of region features in the high-resolution branch.

\subsubsection{High-resolution branch}
Since ViT~\cite{ViT} has achieved satisfactory results in image classification task, more and more transformer-based methods have been proposed for visual feature learning. In our model, the frozen Swin-L Transformer~\cite{Swin_transformer} model is applied to learn global contextual representations from the high-resolution inputs. Based on the shifted window principle, Swin Transformer is a hierarchical four-stage feature extractor, reducing the input resolution by 32 times in total. In each stage, the feature map produced by Swin Transformer has the same downsampling rate as that of ResNet-101, but with fewer number of channels. In the same way, a linear layer and a query former are added to adjust the number and dimension of output features from the high-resolution branch. Here, three query formers in the visual encoders have the same structure and hyper-parameter setting. During the training process, their parameters are trained independently and not shared. 

\subsubsection{Gate-attention module}
To integrate the grid and region features in the low-resolution and high-resolution branch, a gate-attention module is designed. In this module, the grid features from low-resolution branch $L_g \in \mathbb{R}^{Q \times C}$ and the grid features from high-resolution branch $H_g \in \mathbb{R}^{Q \times C}$ are first concatenated in the channel dimension $C_g = [L_g; H_g] \in \mathbb{R}^{Q \times 2C}$, where $Q$ is the query length of the query former and $C$ is the number of feature channels. After the object-level features $L_r$ are enhanced through a self-attention mechanism, $V_r = SA(L_r)$ is gained. Then, the concatenated features and enhanced region features are used as query, key and value in the following cross-attention module, $V = CA(V_r, C_g, C_g)$. 

Considering the possibility redundancy in features extracted from multiple feature branches, a zero-initialized gating factor $w$ is designed to weight the outputs from the attention module and a residual connection operation is added to remain more object-level features: $\hat{V} = w V + L_r$. Besides, the visual features from the gate-attention module need to be inserted into the LLM for caption generation, thus a linear projection operation is added to adjust the dimension from 512-dim to 4096-dim, fitting the caption generator.

\subsection{Caption reasoning module}
\label{caption reasoning module}
In our work, the pre-trained LLaMA-2-7B model is fine-tuned to produce four descriptions for the image including scenario type, risk objects actions, driving intentions and suggestions of the ego-vehicle. The inputs of the LLM consist of two parts: the embedded text prompts and the fused visual features. The prompts designed in our model are tokenized by SentencePiece~\cite{Sentencepiece} method. Then, the pre-trained embedding layer in LLaMA-2-7B is utilized to obtain prompt sequence $T \in \mathbb{R}^{N_l \times D_l}$, where $N_l$ is the text length and $D_l$ is the embedding dimension.

In the caption reasoning module, the prompt sequences are first encoded using a vanilla transformer block, and then multiple transformer blocks with adapters are utilized to integrate visual and linguistic inputs. To better align visual and textual features in the LLM, each learnable adapter is added with visual outputs from the gate-attention module. 

\subsection{Regression module}
\begin{figure}[!t]
	\centering
	\includegraphics[width=0.8\linewidth]{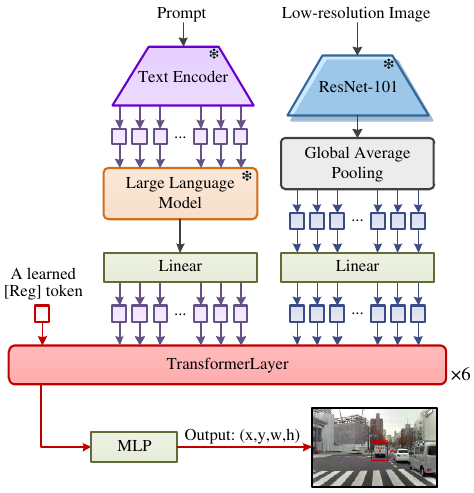}
	\caption{The structure of the transformer-based regression module. (Purple blocks represent text sequences, and blue blocks represent image sequences.)}
	\label{Fig_regression_module}
\end{figure}

\label{localization module}
In the proposed regression module, the generated descriptions and grid features from the low-resolution branch are both utilized to locate the risk object. Here, a learnable regression query $q \in \mathbb{R}^{1 \times C}$ is appended, which is randomly initialized and optimized together with regression network. We denote the visual grid features from ResNet-101 in the low-resolution branch and predicted linguistic tokens as $L_g^{\prime} \in \mathbb{R}^{N_g \times C}$ and $T_l \in \mathbb{R}^{N_l \times C}$. And the regression query is concatenated with visual and linguistic tokens, obtaining $x_0 \in \mathbb{R}^{(1+N_g+N_l) \times C}$. Then a cascaded six-layer transformer network with positional embedding is applied to learn regression coordinates. The specific structure of the regression module is shown in Fig.~\ref{Fig_regression_module}.

For the convenience of model training, we normalize the coordinate values. At the output layer of the transformer, the value at the first position represents the regression coordinates, which are further projected by a MLP and Sigmoid layer. Here, the MLP reduces the output dimension from 512-dim to 4-dim, and the Sigmoid function unifies the range of coordinates. Typically, the output state of the regression query is enriched with visual and linguistic information, containing more precision coordinate information. The training of the regression module involves two loss functions: Smooth L1 loss and generalized IoU loss~\cite{giou_loss} (GIoU loss). Here, we denote the predicted box and the ground-truth box as $\hat{b} = (\hat{x},\hat{y},\hat{w},\hat{h})$ and $b = (x,y,w,h)$, therefore, the total regression loss is:
\begin{equation}
	L = L_{smooth-L1}(b,\hat{b}) + L_{giou}(b,\hat{b}), \\
	\label{Eqa_regression_loss}
\end{equation}
\noindent where $L_{smooth-L1}$ and $L_{giou}$ represent the Smooth L1 loss and GIoU loss, respectively.

\section{EXPERIMENTS}
\subsection{Implementation details}
\subsubsection{Dataset} 

DRAMA-ROLISP~\cite{HiLM-D} is a large-scale risk object description and localization dataset in driving scenes. The dataset is collected in Tokyo, Japan, and contains 11872 training scenes, 2544 validation scenes, and 2544 test scenes, and is an extension of the DRAMA~\cite{drama} dataset released in year 2023. Each image has a resolution of 1928$\times$1280 or 2704$\times$1520. In terms of scene descriptions, DRAMA-ROLISP dataset provides descriptions about the risk object, and driving intentions and suggestions of the ego-vehicle. 

\begin{figure}[!t]
	\centering
	\includegraphics[width=1.0\linewidth]{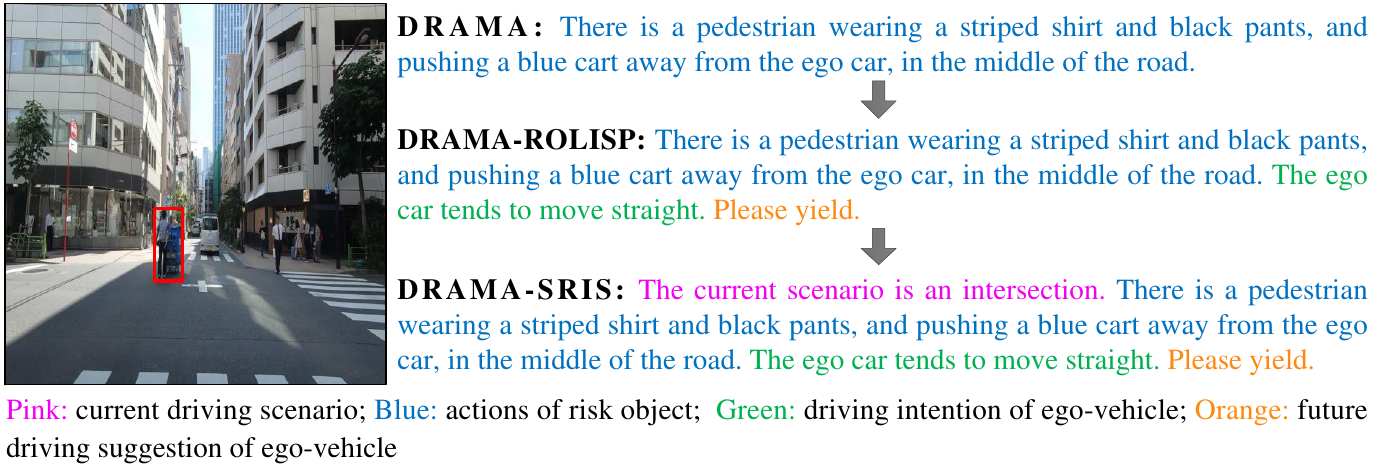}
	\caption{Comparison of text descriptions for three traffic scenario dataset. The coordinate boxes for risk object detection in three datasets are the same.}
	\label{Fig_dataset_compare}
\end{figure}

However, this dataset still lacks a description of the current driving scenario. Therefore, we further extend the descriptions by manually labeling the current driving scenario. Here, we divide all driving scenes in the dataset into three types: conventional urban roads, narrow roads, and intersections, with each image labeled as one of these scenarios. Fig.~\ref{Fig_dataset_compare} compares the descriptions in three datasets, and the descriptions become increasingly comprehensive and rich.

\subsubsection{Training details}
In this paper, our model is implemented under the PyTorch framework using one NVIDIA A800 GPU. The overall training process contains two stages, jointly accomplishing the tasks of scene description and risk object detection. In the first training stage, only the vision encoder and caption reasoning module are trained, and the parameters are optimized through the cross-entropy loss. In the dual-branch visual feature extractor, the input images are resized and cropped to 224$\times$224 and 384$\times$384 used in the low-resolution and high-resolution branch. During the model training, the initial learning rate is set to 5e-4, which is decreased by half for every three epochs. 


In the second training stage, the parameters pre-trained in the first stage are frozen to locate risk objects, only training the regression module from scratch. In this module, each self-attention layer has 8 attention heads with a hidden size of 512. During the model training, the Adam optimizer is used with a smaller fixed learning rate of 1e-4. 


\subsubsection{Evaluation metrics}
Our method comprises two training tasks: (1) describing the driving scenario and risk object while predicting the driving intentions and suggestions of ego-vehicle; (2) predicting the coordinates of risk objects in the scene. The description performance is evaluated using the standard metrics including BLEU-1 (B1)~\cite{B}, BLEU-4 (B4)~\cite{B}, METEOR (M)~\cite{M}, and CIDEr (C)~\cite{C}. And the detection performance is evaluated using the mean intersection over union (mIoU) metric. 


\subsection{Ablation studies}

\begin{table}[!t]
	\renewcommand{\arraystretch}{1.1}
	\caption{Ablation study for different modules in the dual-branch visual feature extractor. ``w/o'' indicates our model without the specific module, ``LR'' indicates the low-resolution branch, ``HR'' indicates the high-resolution branch, ``Gate-atten'' indicates the gate-attention module. ``AVG'' indicates the average of B4 and mIoU. (\%)}
	\centering
	\begin{tabular}{@{}m{0.2cm}<{\centering}|m{1.2cm}|m{0.3cm}<{\centering}m{0.3cm}<{\centering}m{0.6cm}<{\centering}|m{0.3cm}<{\centering}m{1.5cm}<{\centering}|m{0.4cm}<{\centering}}
		\hline
		\multirow{2}*{} & \multirow{2}*{Method} & \multicolumn{3}{c|}{\emph{Image Captioning}} & \multicolumn{2}{c|}{\emph{Localization}} & \multirow{2}*{AVG}  \\
		&  &B4  &M  &C  &mIoU  &Acc(IoU$>$0.5)  &  \\
		\hline
		(a) & w/o LR   &60.4  &43.7  &239.9  &46.7  &58.8  &53.6   \\
		(b) & w/o HR   &60.7  &44.2  &246.6  &57.2  &62.9  &59.0    \\
		(c) & w/o Gate-atten  &63.5  &45.1  &279.2  &50.9  &60.0   &57.2    \\
		(d) &\textbf{MLLM-SUL}  &\textbf{65.2}  &\textbf{45.7}  &\textbf{298.5}  &\textbf{59.6}  &\textbf{64.4}   &\textbf{62.4}    \\
		\hline
	\end{tabular}
	\label{Table_ablation_study_visual}
\end{table}

\subsubsection{Different components of dual-branch visual encoder}

In this section, we conduct ablation studies on the DRAMA-SRIS dataset to demonstrate the effectiveness of each visual encoder branch. The results are reported in Table~\ref{Table_ablation_study_visual}, and four models are designed: (a) only a high-resolution branch is presented in the visual encoder without region features; (b) only a low-resolution branch in the visual encoder, and the gate-attention module is utilized to fuse the grid and region features from ResNet-101 and object detector; (c) a dual-resolution branch is performed in the visual encoder, and multiple outputs are simply combined through a simple concatenation operation; (d) both dual-resolution branch and gate-attention module are implemented in the encoder. In the regression module, except for visual inputs in method (a) are from Swin Transformer, the other three visual inputs all come from ResNet-101 network. In terms of evaluation metrics, we compare the results of image captioning and detection task, as well as the average accuracy. 

When the visual encoder only contains one resolution branch, the text generation accuracy is fairly low, resulting in an equally low detection precision, \emph{i.e.}, method (a) and (b) only achieve 239.9\% and 246.6\% CIDEr score, far lower than the score of 298.5\% acquired by our method. For the average regression results of all boxes, method (a) and (b) are 12.9\% and 2.4\% lower than ours, respectively. 
These results fully demonstrate the significance of the dual-resolution visual encoder, which are superior to an individual low-resolution or high-resolution branch. Besides, the experimental results of (c) and (d) show the importance of gate-attention module. when the gate-attention module is replaced with a simple concatenation operation, the captioning BLEU-4 and detection mIoU score decrease by 1.7\% and 8.7\%, respectively.

\subsubsection{Different visual inputs of regression module}
\begin{table}[!t]
	\renewcommand{\arraystretch}{1.1}
	\caption{Ablation study for different visual inputs in the regression module. ``$\mathcal{D}$'', ``$\mathcal{S}$'' and ``$\mathcal{R}$'' indicate region features from the detector, grid features from Swin Transformer, and grid features from ResNet-101. And $[;]$ indicates concatenating two features. ``AVG'' indicates the average of B4 and mIoU. (\%)}
	\centering
	\begin{tabular}{@{}m{0.2cm}<{\centering}|m{1.4cm}<{\centering}|m{0.3cm}<{\centering}m{0.3cm}<{\centering}m{0.6cm}<{\centering}|m{0.4cm}<{\centering}m{1.5cm}<{\centering}|m{0.5cm}<{\centering}}
		\hline
		\multirow{2}*{} & \multirow{2}*{Visual inputs} & \multicolumn{3}{c|}{\emph{Image Captioning}} & \multicolumn{2}{c|}{\emph{Localization}} & \multirow{2}*{AVG}  \\
		& &B4  &M  &C  &mIoU  &Acc(IoU$>$0.5)  &  \\
		\hline
		(a) & $\mathcal{D}$  &65.2  &45.7  &298.5  &50.0 &59.6   &57.6   \\
		(b) & $\mathcal{S}$  &65.2  &45.7  &298.5  &54.2 &61.9   &59.7   \\
		(c) & [$\mathcal{S}$; $\mathcal{R}$]   &65.2  &45.7  &298.5  &57.9 &64.0   &61.6   \\
		(d) & $\mathcal{R}$   &65.2  &45.7  &298.5  &\textbf{59.6}  &\textbf{64.4}   &\textbf{62.4}    \\
		\hline
	\end{tabular}
	\label{Table_ablation_study_regression}
\end{table}

In this section, three different visual inputs are compared in the regression module: (a) only region features from the Faster R-CNN detector; (b) only grid features from Swin-Transformer; (c) grid features from Swin-Transformer and ResNet-101 are concatenated; (d) only grid features from ResNet-101. The captioning and localization accuracy of each method are reported in Table~\ref{Table_ablation_study_regression}. Among four methods, the network structure of the image captioning task is the same, so they have the same text prediction accuracy. 

When region features from Faster R-CNN or grid features from Swin-Transformer are employed in the regression module, the average localization accuracy are only 50.0\% and 54.2\%. While using the grid features from the low-resolution branch, the localization precision can reach the highest 59.6\%. Furthermore, we concatenate two grid features in the dimension of sequence length, resulting in a lower localization accuracy of 57.9\%. From above experimental results, the grid features from ResNet-101 are most suitable for inputting into the regression network to locate risk objects. 

\subsubsection{Different backbone networks in the high-resolution branch}
\begin{table}[!t]
	\renewcommand{\arraystretch}{1.1}
	\caption{Ablation study for different backbone networks in the high-resolution branch. ``AVG'' indicates the average of B4 and mIoU. (\%)}
	\centering
	\begin{tabular}{@{}m{0.3cm}<{\centering}|m{2.0cm}<{\centering}|m{0.4cm}<{\centering}m{0.4cm}<{\centering}m{0.8cm}<{\centering}|m{1.4cm}<{\centering}|m{0.5cm}<{\centering}}
		\hline
		\multirow{2}*{} &\multirow{2}*{Model}  
		&\multicolumn{3}{c|}{\emph{Image Captioning}} &\emph{Localization} & \multirow{2}*{AVG}  \\
		&  &B4  &M  &C  &mIoU  &  \\
		\hline
		(a) &SwinV2-S~\cite{SwinV2_transformer}  &64.8 &45.6  &289.0  &56.8   &60.8    \\
		(b) &SwinV2-B~\cite{SwinV2_transformer}  &64.0 &45.2  &273.4  &56.6   &60.3    \\
		(c) &SwinMLP-T~\cite{Swin_transformer}  &65.1 &45.7  &288.0  &55.7   &60.4    \\
		(d) &Swin-B~\cite{Swin_transformer}    &60.6 &44.0  &225.1  &56.1   &58.4    \\
		(e) & Swin-L~\cite{Swin_transformer}   &\textbf{65.2}  &\textbf{45.7}  &\textbf{298.5}  &\textbf{59.6}  &\textbf{62.4}    \\
		\hline
	\end{tabular}
	\label{Table_ablation_study_backbone_high_resolution}
\end{table}

\begin{table*}[!t]
	\renewcommand{\arraystretch}{1.1}
	\caption{Comparison with state-of-the-art methods on the DRAMA-ROLISP dataset. ``IoU$_{\emph{S}}$" indicates the IoU of small-size objects, ``IoU$_{\emph{M}}$" indicates the IoU of medium-size objects, ``IoU$_{\emph{L}}$" indicates the IoU of large-size objects, and ``AVG'' indicates the average of B4 and mIoU score. (\%)}
	\centering
	\begin{tabular}{@{}m{4.5cm}|m{1.2cm}<{\centering}m{1.2cm}<{\centering}m{1.2cm}<{\centering}|m{1.1cm}<{\centering}m{1.1cm}<{\centering}m{1.1cm}<{\centering}m{1.1cm}<{\centering}|m{1.2cm}<{\centering}}
		\hline
		\multirow{2}*{Method} & \multicolumn{3}{c|}{\emph{Image Captioning}} & \multicolumn{4}{c|}{\emph{Localization}} & \multirow{2}*{AVG}  \\
		&B4  &M  &C  &mIoU  &IoU$_{\emph{S}}$ &IoU$_{\emph{M}}$ &IoU$_{\emph{L}}$  &  \\
		\hline
		BLIP-2~\cite{blip2}   &46.1  &34.3  &194.7  &46.3  &8.1  &60.2  &73.7  &46.2   \\
		eP-ALM~\cite{ep-alm}   &51.4  &38.0  &225.1  &43.2  &7.2  &56.8  &68.8  &47.3   \\
		LLaVA~\cite{LLaVA}  &47.5  &35.2   &198.6  &47.2  &8.0  &62.1  &74.2  &47.4   \\
		Video-LLaMA~\cite{video-llama}   &53.9  &37.8  &229.5  &42.8  &6.9  &55.3  &67.9  &48.4   \\
		InstructBLIP~\cite{InstructBLIP}  &49.9  &37.9   &205.0  &47.8  &9.1  &62.2  &74.5  &48.9   \\
		Shikra~\cite{shikra}  &49.8  &37.7  &204.7  &50.3  &10.4  &59.5  &73.8 &50.1   \\
		LLaMA-Guard-3-8B + Adapter~\cite{llama-adapter}  &56.5   &41.8  &261.0   &52.3  &48.9  &55.3  &55.9 &54.4   \\
		LLaMA-3.1-8B-Instruct + Adapter~\cite{llama-adapter}  &55.8   &41.1  &258.7   &53.0  &\textbf{49.6}  &55.9  &56.6 &54.4   \\
		LLaMA-3.2-3B-Instruct + Adapter~\cite{llama-adapter}  &57.9   &42.2  &275.9   &53.0  &49.5  &55.9  &56.7 &55.5   \\
		HiLM-D~\cite{HiLM-D}  &58.5  &41.2   &\textbf{279.2}  &59.2  &31.1  &62.5  &82.3   &58.9   \\
		\textbf{MLLM-SUL}   &\textbf{60.1}  &\textbf{43.7} &276.1  &\textbf{59.3}  &36.5  &\textbf{72.1}  &\textbf{86.4}  &\textbf{59.7}   \\
		\hline
	\end{tabular}
	\label{Table_drama_rolisp}
\end{table*}

\begin{table}[!t]
	\renewcommand{\arraystretch}{1.1}
	\caption{Comparison results between our method and HiLM-D on the DRAMA-ROLISP dataset. FLOPs and memory are calculated with a batch size of one.}
	\centering
	\begin{tabular}{@{}m{1.6cm}|m{1.0cm}<{\centering}m{1.0cm}<{\centering}m{1.2cm}<{\centering}m{1.5cm}<{\centering}}
		\hline
		Method  & Memory (GB) $\downarrow$  & FLOPs (10$^{10}$) $\downarrow$ & Captioning B4 (\%) $\uparrow$  & Localization mIoU (\%) $\uparrow$   \\
		\hline
		HiLM-D~\cite{HiLM-D}  &28.2  &159.5  &58.5  &59.2 \\
		\textbf{MLLM-SUL}  &\textbf{26.4}  &\textbf{67.2}  &\textbf{60.1}  &\textbf{59.3} \\ 
		\hline
	\end{tabular}
	\label{Table_compare_hilmd_ours}
\end{table}

In this section, we explore four backbone networks in the high-resolution branch and the results are reported in Table~\ref{Table_ablation_study_backbone_high_resolution}: (a) SwinV2-S~\cite{SwinV2_transformer} model with 256$\times$256 input; (b) SwinV2-B~\cite{SwinV2_transformer} model with 256$\times$256 input; (c) SwinMLP-T~\cite{Swin_transformer} model with 256$\times$256 input; (d) Swin-B~\cite{Swin_transformer} model with 384$\times$384 input; (e) Swin-L~\cite{Swin_transformer} model with 384$\times$384 input. The results of four ablation studies indicate that when Swin-B is used as the backbone in the high-resolution branch, the model has the lowest text prediction accuracy and localization precision, with a BLEU-4 score of only 60.6\% and mIoU value of only 56.1\%. Taking into account the performance of two tasks, Swin-L network with an input resolution of 384$\times$384 is the most suitable as the backbone of the high-resolution visual branch.

\subsection{Comparison with state-of-the-art MLLMs} 


\begin{table*}[!t]
	\renewcommand{\arraystretch}{1.1}
	\caption{Comparison with state-of-the-art methods on the DRAMA-SRIS dataset. ``AVG'' indicates the average of B4 and mIoU. (\%)}
	\centering
	\begin{tabular}{@{}m{4.9cm}|m{1.2cm}<{\centering}m{1.2cm}<{\centering}m{1.2cm}<{\centering}m{1.3cm}<{\centering}|m{1.2cm}<{\centering}m{2.0cm}<{\centering}|m{1.4cm}<{\centering}}
		\hline
		\multirow{2}*{Method} & \multicolumn{4}{c|}{\emph{Image Captioning}} & \multicolumn{2}{c|}{\emph{Localization}} & \multirow{2}*{AVG}  \\
		&B1  &B4  &M  &C  &mIoU  &Acc(IoU$>$0.5)  &  \\
		\hline
		BLIP-2 OPT$_{\mathrm{2.7B}}$~\cite{blip2}  &76.1  &60.8  &41.9 &208.3  &53.0    &61.3  &56.9   \\
		BLIP-2 OPT$_{\mathrm{6.7B}}$~\cite{blip2}  &76.3  &61.1  &42.1 &208.9  &53.5    &61.5  &57.3   \\
		InstructBLIP OPT$_{\mathrm{2.7B}}$~\cite{InstructBLIP}  &74.7  &57.7  &40.1 &197.5  &55.7   &62.9  &56.7   \\
		InstructBLIP OPT$_{\mathrm{6.7B}}$~\cite{InstructBLIP}  &74.8  &57.6  &40.1   &193.5  &55.7    &62.8  &56.7   \\
		LLaMA-2-7B + Adapter~\cite{llama-adapter}  &78.7  &63.4  &44.5  &275.0  & 57.8   &63.7  &60.6   \\
		LLaMA-3.1-8B + Adapter~\cite{llama-adapter}  &71.5   &56.5   &39.0   &263.3   &52.9    &61.8   &54.7    \\
		LLaMA-3.1-8B-Instruct + Adapter~\cite{llama-adapter}  &71.0   &56.7   &39.0   &265.3    &53.0   &61.6   &54.9    \\
		LLaMA-3.2-3B + Adapter~\cite{llama-adapter}  &71.5   &56.8   &39.2   &263.8   &52.7    &61.4   &54.8    \\
		LLaMA-3.2-3B-Instruct + Adapter~\cite{llama-adapter}   &72.2   &57.4   &39.6   &272.4   &51.9    &61.1   &54.7     \\
		LLaMA-Guard-3-8B + Adapter~\cite{llama-adapter}  &72.0   &56.7   &39.2   &263.2   &53.2    &61.5   &55.0    \\
		\textbf{MLLM-SUL}  &\textbf{80.1}  &\textbf{65.2}  &\textbf{45.7}  &\textbf{298.5}  &\textbf{59.6}  &\textbf{64.4}  &\textbf{62.4}   \\
		\hline
	\end{tabular}
	\label{Table_drama_sris}
\end{table*}

\subsubsection{Results on the DRAMA-ROLISP dataset} We compare MLLM-SUL with several image-based and video-based MLLMs proposed in recent years, including BLIP-2~\cite{blip2}, eP-ALM~\cite{ep-alm}, LLaVA~\cite{LLaVA}, Video-LLaMA~\cite{video-llama}, InstructBLIP~\cite{InstructBLIP}, Shikra~\cite{shikra}, HiLM-D~\cite{HiLM-D}, and several LLaMA-based methods. 
The comparison results reported in Table~\ref{Table_drama_rolisp} demonstrate the superior performance of our model in both captioning and localization tasks. Among these models, BLIP-2, LLaVA, InstructBLIP, Shikra, and LLaMA+adapter methods are image-based methods, which all have lower captioning accuracy and localization precision than ours. For example, BLIP-2 has a 14.0\% lower BLEU-4 score than ours, LLaVA has a 12.6\% lower BLEU-4 score than ours, and LLaMA-3.1-8B-Instruct+Adapter has a 17.4\% lower CIDEr score than ours.
Although our method is image-based, it outperforms video-based methods eP-ALM and Video-LLaMA in both captioning and localization metric, \emph{i.e.}, MLLM-SUL improves the BLEU-4 score by 8.7\% compared with eP-ALM, by 6.2\% compared with Video-LLaMA, and by 1.6\% compared with HiLM-D. And while considering the regression task, our model also has the highest localization accuracy for objects with different sizes.

Furthermore, we compare MLLM-SUL with HiLM-D in terms of model memory, floating-point operations (FLOPs), captioning accuracy, and localization precision in Table~\ref{Table_compare_hilmd_ours}. \textcolor{black}{The main differences between HiLM-D and our method lie in three aspects. First, a high-precision object detector is added in our method to extract object-level features. Second, HiLM-D uses the GradCAM module to assist in locating high-risk objects, while our model uses a much simpler regression query to learn location information. Third, the input of the low-resolution branch is the current frame image in our method, rather than a video of frames in the HiLM-D. This eliminates the learning of temporal data associations, reducing computational complexity.} \textcolor{black}{The comparison results demonstrate that the object-level features in our model help improve the captioning performance, with a 1.6\% increase in BLEU-4 score. Meanwhile, our model reduced the memory and FLOPs of HiLM-D by 6.4\% and 57.9\%, respectively. Thus, MLLM-SUL achieves more superior performance with fewer operations and computational resources.
}


\begin{figure*}[!t]
	\centering
	\includegraphics[width=1.0\linewidth]{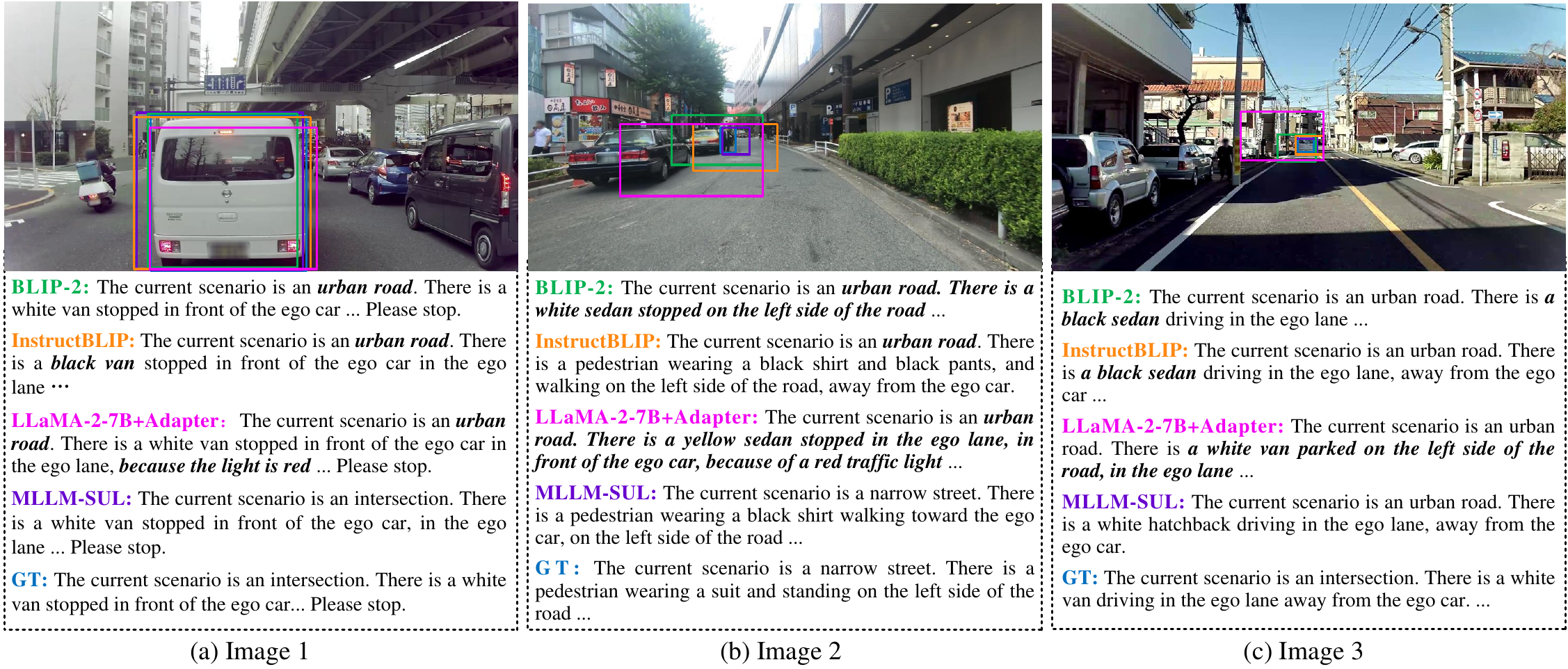}
	\caption{Qualitative results in the DRAMA-SRIS dataset. The language generator in BLIP-2 and InstructBLIP is OPT$_{\mathrm{6.7B}}$. ``GT" indicates the ground-truth results and wrong predictions are represented in italics. For the same MLLM, descriptions and bounding box are represented by the same color.}
	\label{Fig_compare_results}
\end{figure*}



\subsubsection{Results on the DRAMA-SRIS dataset} In the extended DRAMA-SRIS dataset, the comparison results are reported in Table~\ref{Table_drama_sris}. Compared with LLaMA-2-7B+Adapter method, our model improves the BLEU-1 score by 1.4\%, BLEU-4 score by 1.8\%, METEOR score by 1.2\%, and CIDEr score by 23.5\% for captioning task. For the risk object localization, our model achieves the highest mean IoU of 59.6\%. When only counting boxes with IoU greater than 0.5, our method achieves the highest 64.4\% regression accuracy. Compared with the more advanced LLaMA-3.2-3B-Instruct+Adapter, our model improves the CIDEr and mIoU score by 26.1\% and the 7.7\%, respectively.

The visualization results of different MLLMs are shown in Fig.~\ref{Fig_compare_results}. In Fig.~\ref{Fig_compare_results}(a), a white van in front is very close to the -vehicle, with the highest risk in the current driving scenario. InstructBLIP wrongly predicts the color of the van, and the red traffic light generated by LLaMA-2-7B+Adapter cannot be found in the image. While for the large-size object in the image, the localization precision of each method is pretty high. In Fig.~\ref{Fig_compare_results}(b), BLIP-2 and LLaMA-2-7B+Adapter incorrectly predict the risk objects in the current driving scene, and InstructBLIP incorrectly predicts the current scenario as an urban road not a narrow street. Meanwhile, the size of the bounding box generated by three comparison methods is much larger than the ground-truth size, indicating that the proposed MLLM-SUL is more accurate in describing and locating small-size risk objects. Compared with the close-up van in Fig.~\ref{Fig_compare_results}(a), the long-distance and small-size van in Fig.~\ref{Fig_compare_results}(c) is more difficult to describe and locate. Obviously, LLaMA-2-7B+Adapter locates the wrong risk object, while BLIP-2 and InstructBLIP predict the wrong color of the van in the driving scene. And the regression box predicted by the MLLM-SUL is the closest to the ground-truth box, demonstrating that our model has better effect in describing and locating long-distance risk objects.


\section{CONCLUSIONS}
We propose a novel MLLM-SUL method for jointly describing the driving scenario and locating the risk object in traffic scenes. In our model, a visual encoder consisting of a low-resolution and a high-resolution branch is aggregated into LLaMA model, providing more accurate and comprehensive descriptions of scene types, risk object actions, and driving intentions and suggestions. In addition, the design of a dual-resolution encoder with different receptive fields has better feature extraction ability for small-size, medium-size, and large-size objects, improving the localization accuracy for different types of risk objects. The experimental results on DRAMA-ROLISP and DRAMA-SRIS datasets demonstrate that our method outperforms many video-based and image-based MLLMs, achieving state-of-the-art effects for multimodal scene understanding tasks. 


\section*{ACKNOWLEDGMENT}

We would like to thank the anonymous reviewers for their valuable feedback on our work. This research is supported in part by the National Nature Science Foundation of China (Nos. 62373289, 62273256) and in part by the Fundamental Research Funds for the Central Universities.

\bibliographystyle{IEEEtran}
\bibliography{root}

\end{document}